\documentclass{article}
\usepackage{subcaption}

\usepackage{microtype}
\usepackage{graphicx}
\usepackage{booktabs} % for professional tables

\usepackage{hyperref}

\usepackage[accepted]{icml2025}

\usepackage{amsmath}
\usepackage{amssymb}
\usepackage{mathtools}
\usepackage{amsthm} % Use algorithm2e for the required style

\usepackage{graphicx}  % Required for including images
\usepackage{subcaption} % For subfigures
\usepackage{float}
\usepackage{multirow}
\usepackage{svg}
\usepackage[capitalize,noabbrev]{cleveref}
\theoremstyle{plain}

\theoremstyle{definition}

\theoremstyle{remark}

\usepackage[textsize=tiny]{todonotes}

\icmltitlerunning{}

\begin{document}

\twocolumn[
\icmltitle{DiffEx: Explaining a Classifier with Diffusion Models to Identify Microscopic Cellular Variations}



\icmlsetsymbol{equal}{*}

\begin{icmlauthorlist}
\icmlauthor{Anis Bourou}{equal,yyy,comp}
\icmlauthor{Saranga Kingkor Mahanta}{equal,yyy}
\icmlauthor{Thomas Boyer}{yyy}
\icmlauthor{Valérie Mezger}{comp}
\icmlauthor{Auguste Genovesio}{yyy}
\end{icmlauthorlist}

\icmlaffiliation{yyy}{ENS}
\icmlaffiliation{comp}{Université Paris Cité}

\icmlcorrespondingauthor{Auguste Genovesio}{auguste.genovesio@bio.ens.psl.eu}

\icmlkeywords{Machine Learning, ICML}

\vskip 0.3in
]



\printAffiliationsAndNotice{\icmlEqualContribution} % otherwise use the standard text.

\begin{abstract}
In recent years, deep learning models have been extensively applied to biological data across various modalities. Discriminative deep learning models have excelled at classifying images into categories (e.g., healthy versus diseased, treated versus untreated). However, these models are often perceived as black boxes due to their complexity and lack of interpretability, limiting their application in real-world biological contexts. In biological research, explainability is essential: understanding classifier decisions and identifying subtle differences between conditions are critical for elucidating the effects of treatments, disease progression, and biological processes.
To address this challenge, we propose DiffEx, a method for generating visually interpretable attributes to explain classifiers and identify microscopic cellular variations between different conditions. We demonstrate the effectiveness of DiffEx in explaining classifiers trained on natural and biological images. Furthermore, we use DiffEx to uncover phenotypic differences within microscopy datasets. By offering insights into cellular variations through classifier explanations, DiffEx has the potential to advance the understanding of diseases and aid drug discovery by identifying novel biomarkers.
\end{abstract}

\section{Introduction}
\label{submission}

Image classification is a fundamental task in deep learning that has achieved remarkable results~\cite{convnets,resnet,densenet,dosovitskiy2020vit,convnext}. The success of  classifiers is primarily due to their ability to extract patterns and features from images to distinguish between classes. However, these patterns can often be difficult to discern~\cite{convnets, understanding_convnets}, particularly in microscopy images~\cite{microscopy_challenges1,microscopy_challenges2}, which poses challenges for the interpretability of these models. Explaining the decision-making processes of discriminative models is an active area of research. Various strategies~\cite{grad_cam1,grad_cam2,Lang_2021_ICCV,time_WACV, Jeanneret_2023_CVPR} have been proposed to clarify how deep learning models arrive at their outputs, aiming to make the decision processes more transparent and understandable.

In biological imaging, interpreting classifier decisions is essential for understanding extracted features and uncovering biological insights. For instance, when classifying healthy versus diseased tissues or treated versus untreated samples, it is crucial to determine which attributes influence predictions. Identifying these cellular variations—phenotypes—not only deepens our understanding of diseases but also clarifies treatment effects. Thus, pinpointing the attributes that drive classifier outcomes is fundamental. By uncovering them, we can reveal biologically meaningful phenotypes that offer deeper insights into complex phenomena~\cite{bourou,phenexp,bourou_2}.

In this work, we introduce \textbf{DiffEx}, a method for uncovering the attributes leveraged by a classifier to make its decisions, and demonstrate its effectiveness on both natural and microscopy images. Our method first builds a latent space that incorporates the classifier's attributes using diffusion models. We then identify interpretable directions in this latent space using a contrastive learning approach. The discovered directions are ranked by selecting those that most significantly change the classifier's decision.

\noindent We summarize our contributions in this work as follows:

\begin{itemize}
\item We introduce DiffEx, a novel method leveraging diffusion models to identify interpretable attributes that explain the decisions of a classifier. 

\item We demonstrate the versatility of DiffEx by applying it to classifiers trained on both natural and biological image datasets. 

\item In biological datasets, we employ DiffEx to uncover subtle cellular variations between different conditions.
\end{itemize}

\section{Related Work}

\subsection{Classifiers Explainability}
Class Activation Maps (CAMs)~\cite{grad_cam1,grad_cam2} are a well-known technique for explaining classifier decisions, as they highlight the most influential regions in an image that affect the classifier's output. However, these methods typically require access to the classifier's architecture and all its layers, as they involve computing gradients of the outputs with respect to the inputs. Additionally, CAMs only indicate important regions in images without explicitly identifying the affected attributes, such as shape, color, or size. This can be limiting, particularly in microscopy images where subtle variations are of interest. Counterfactual visual explanations represent another family of methods aimed at explaining classifier decisions. These methods seek to identify minimal changes that would alter the classifier's decision. Generative models have been widely used to generate such counterfactual explanations. Generative Adversarial Networks (GANs), for instance, have been employed for this purpose~\cite{Singla2020,Lang_2021_ICCV,Goetschalckx_2019_ICCV_ganalyze}. While some approaches generate counterfactual explanations all at once~\cite{Singla2020,Goetschalckx_2019_ICCV_ganalyze}, the work in~\cite{Lang_2021_ICCV} identifies a set of attributes that influence the classifier's decision. However, GANs suffer from training instability due to the simultaneous optimization of two networks: the generator and the discriminator. Recently, diffusion models have demonstrated more stable training, superior generation quality, and greater diversity~\cite{diffusion_beat_gans,improved_denoising_diffusion_models}. They have also been adopted for generating visual counterfactual explanations~\cite{dvce,time_WACV,global_counterfactual_explanations}.

\subsection{Diffusion Models} Generative models have recently achieved significant success in various tasks~\cite{gan,score_based_generative_models,diffusion_beat_gans,vae}. Diffusion models~\cite{ddpm,ddim,diffusion_beat_gans,improved_denoising_diffusion_models}, a class of generative models, have been applied to different domain~\cite{diffusion_beat_gans,Diffusion_Bioinformatics,ldm}. These models consist of two processes: a known forward process that gradually adds noise to the input data, and a learned backward process that iteratively denoises the noised input. Numerous works have proposed improvements to diffusion models~\cite{diffusion_beat_gans,ldm,improved_denoising_diffusion_models}, enhancing their performance and making them the new state-of-the-art in generative modeling across different tasks. Recently, it has been shown that diffusion models can be used to learn meaningful representations of images that facilitate image editing tasks~\cite{diffusion_ae,diffusion_semantic_space}. In~\cite{diffusion_ae}, the authors proposed adding an encoder network during the training of diffusion models to learn a semantic representation of the image space. This approach enables the model to capture high-level features that can be manipulated for various applications. In~\cite{diffusion_semantic_space}, the authors modified the reverse process—introducing an asymmetric reverse process—to discover semantic latent directions in the space induced by the bottleneck of the U-Net~\cite{Unet} used as a denoiser in the diffusion model, which they refer to as the \textit{h-space}. By exploring this space, they were able to identify directions corresponding to specific semantic attributes, allowing for targeted image modifications. These advancements demonstrate the potential of diffusion models not only for high-quality data generation but also for learning rich representations that can be leveraged for downstream tasks.
\subsection{Detecting phenotypes in microscopy images}
Capturing the visual cellular differences in microscopy images under varying conditions is essential for understanding certain diseases and the effects of treatments~\cite{cellular_profiling,cellular_profiling_2,cellular_profiling_3, bourou,phenexp,bourou_2}. Historically, hand-crafted methods were employed to measure changes between different conditions~\cite{cellprofiler}. However, these tools have limitations, especially when the observed changes are subtle or masked by biological variability~\cite{phenexp,bourou_2}. Recently, generative models have been proposed to alleviate these limitations. In~\cite{bourou}, CycleGAN~\cite{cyclegan} was used to perform image-to-image translations, aiming to discard biological variability and retain only the induced changes. By translating images from one condition to another, the model focused on the specific alterations caused by the experimental conditions, effectively highlighting phenotypic differences. In~\cite{phenexp}, a conditional StyleGAN2~\cite{stylegan2} was trained to identify phenotypes by interpolating between classes in the StyleGAN's latent space. This approach enabled the generation of high-fidelity images that represent different phenotypic expressions, facilitating the study of subtle cellular variations and providing insights into the underlying biological processes. Furthermore, recent advancements have seen the use of conditional diffusion models in image-to-image translation~\cite{bourou_2}. In this method, an image from the source condition is first inverted into a latent code, that is used to generate corresponding the image from the target condition. This technique leverages the strengths of diffusion models in capturing complex data distributions and performing realistic translations between conditions. All of these methods have proven effective in uncovering phenotypes and enhancing the understanding of cellular differences. However, they rely solely on generative models and do not integrate classifiers that can extract patterns from images and assess how a given image would be transferred to another class. Incorporating discriminative models alongside generative approaches could enhance pattern recognition and provide a more comprehensive analysis of cellular changes, ultimately improving the assessment of disease progression and treatment effects.

\begin{figure*}[ht]
  \centering
  \includegraphics[width=0.9\linewidth]{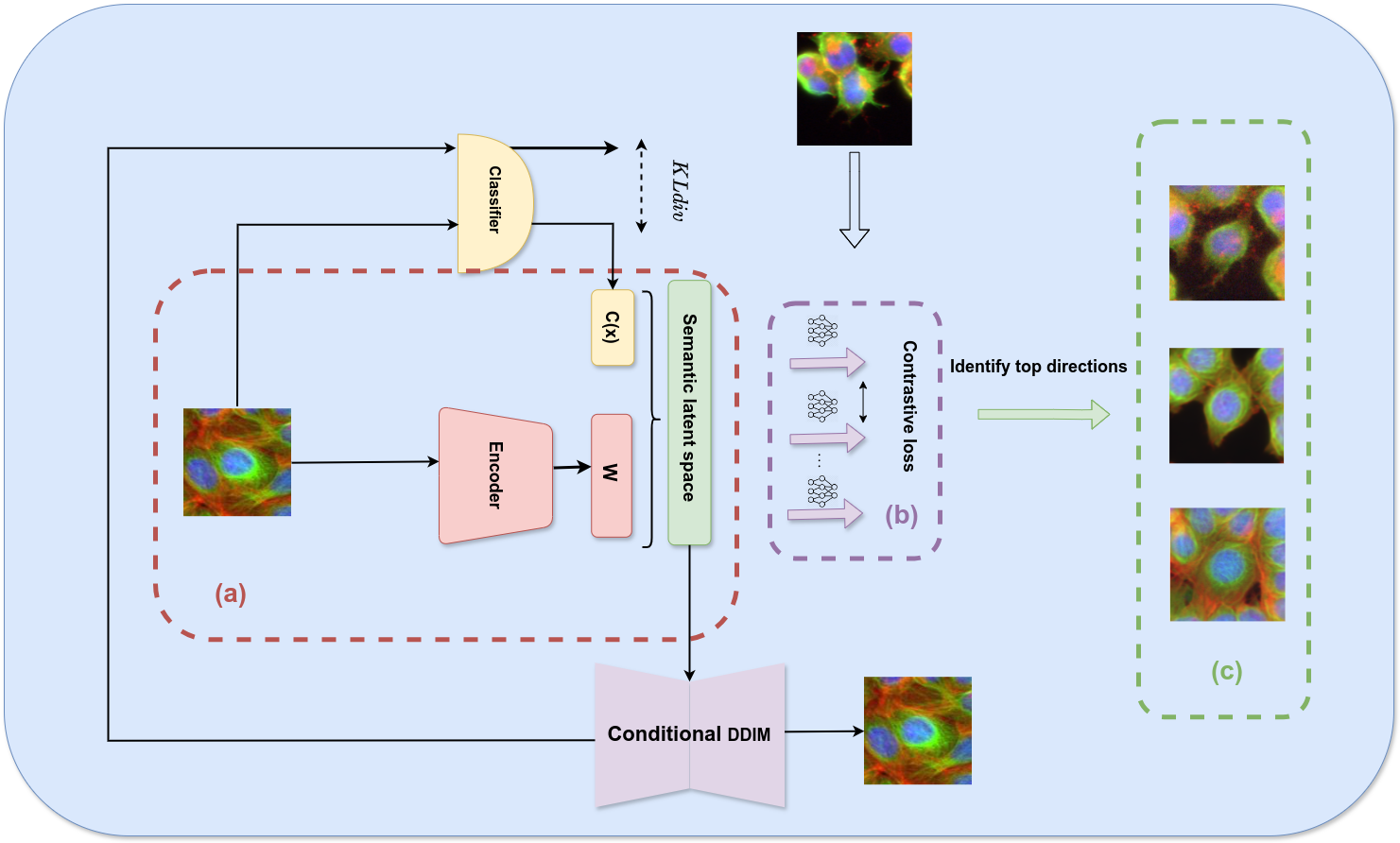} % Replace with your image file name
  \caption{DiffEx primarily consists of three stages: \textbf{(a)} A semantic latent space is constructed by combining the embedding obtained from an encoder with the classifier's prediction for each image. The resulting representation is used to condition the DDIM. \textbf{(b)} Directional models are learned in this semantic latent space using a self-supervised approach. \textbf{(c)} After identifying the directions that most significantly affect the classification probability, we shift the images accordingly. For example, in the accompanying figure, a single image is shifted along the identified directions, resulting in visibly different images that highlight the changes induced by these directions.}
  \label{fig:main_figure}
\end{figure*}

\subsection{Contrastive learning}

Contrastive learning is a powerful self-supervised framework that has achieved remarkable success across various domains, including computer vision and natural language processing~\cite{simclr,clip,language_1,language_2}. By contrasting positive and negative pairs, it learns rich feature representations, maximizing similarity for positive pairs while minimizing it for negative ones using a contrastive loss~\cite{simclr,INFO_NCE,triplet_loss,contrastive_learning}. This versatile approach has been integrated into diverse architectures, enabling the extraction of robust and generalizable features for a wide range of downstream tasks. Beyond traditional applications, contrastive learning has also been leveraged in generative modeling. It has been employed to enhance conditioning in GANs~\cite{contragan} and to improve style transfer in diffusion models~\cite{diffusion_contra}. Discovering interpretable directions in generative models is fundamental to various image generation and editing tasks~\cite{latent_clr,noise_clr,diffusion_semantic_space}. In this context, contrastive learning has proven highly effective. For instance, LatentCLR~\cite{latent_clr} identifies meaningful transformations by applying contrastive learning to the latent space of GANs, while NoiseCLR~\cite{noise_clr} uncovers semantic directions in pre-trained text-to-image diffusion models like Stable Diffusion~\cite{stable_diffusion}.

\section{Method}
In this section, we introduce DiffEx, a method designed to explain a classifier by generating separable and interpretable attributes. As illustrated in Fig~\ref{fig:main_figure}, our method leverages diffusion models to provide insights into the classifier's behavior. First, we construct a latent semantic space that is aware of the classifier specific attributes. Then, using a contrastive learning approach, we identify separable and interpretable directions within this space. Finally, we rank the importance of the discovered directions and modify the image accordingly to highlight the critical features influencing the classifier’s predictions. 
\subsection{Building a classifier-aware semantic latent space}
GANs benefit from a well-structured semantic latent space, which allows for easy control over different attributes of generated samples~\cite{stylegan,stylegan2,biggan,voynov2020unsupervised}. This property has been leveraged in various applications, such as counterfactual visual explanations~\cite{explaining_in_style}. However, due to the iterative nature of diffusion models, they lack such a readily accessible latent space. In this work, we follow an approach similar to~\cite{diffusion_ae}, where we construct a semantic latent space for our diffusion model by incorporating an encoder network. The encoder generates a latent code from a given input image, which is subsequently used to condition the diffusion process. To ensure that the generated samples maintain classifier-relevant attributes, we concatenate the classification score with the latent vector, forming a semantic code to condition the diffusion model, we denote it as $z_{sem}$. 
\begin{equation}
L_{\text{diffusion}} = \sum_{t=1}^{T} \mathbb{E}_{x_0, \epsilon_t} \left[ \left\| \epsilon_\theta \left( x_t, t, z_{\text{sem}} \right) - \epsilon_t \right\|_2^2 \right]
\end{equation}

\begin{figure*}[ht]
  \centering
  \includegraphics[width=\linewidth]{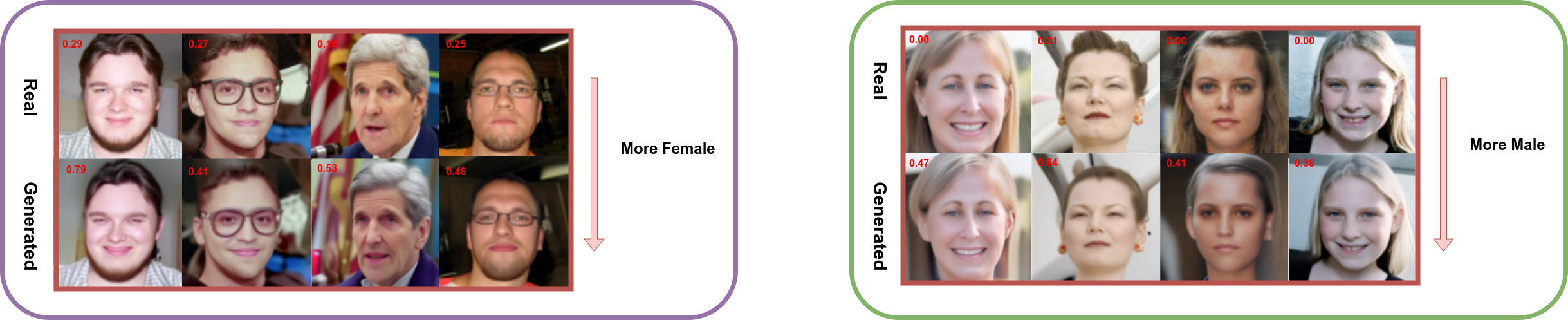} % Replace with your image file name
  \caption{Shifting images toward the opposite class using directions identified by Diffex.
\textbf{Left}: When transforming male images toward the female class, the appearance of lipstick becomes noticeable, suggesting it as a discriminative attribute for the classifier. \textbf{Right}: When shifting female images toward the male class, hairstyles tend to become shorter, indicating an attribute associated with the male class. The probabilities of the target classes are shown in red.}
  \label{fig:figure_faces}
\end{figure*}

\begin{figure}[ht]
  \centering

  % First subfigure
  \begin{subfigure}[b]{0.4\textwidth}  
    \centering
    \includegraphics[width=\textwidth]{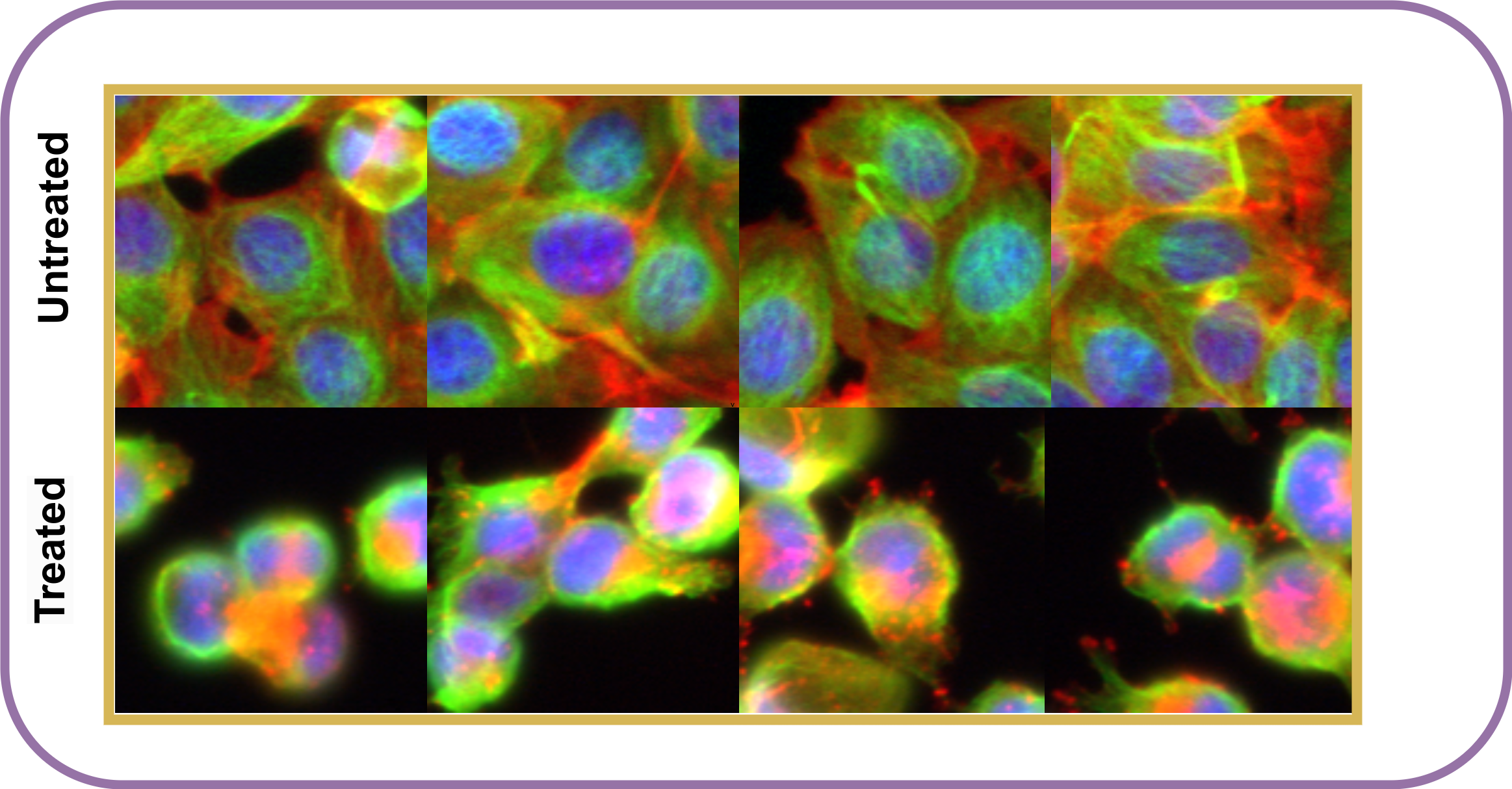} %
    \caption{}
    \label{fig:subfig1}
  \end{subfigure}

  \vspace{1em} % Optional vertical space between subfigures

  % Second subfigure
  \begin{subfigure}[b]{0.4\textwidth}  
    \centering
    \includegraphics[width=\textwidth]{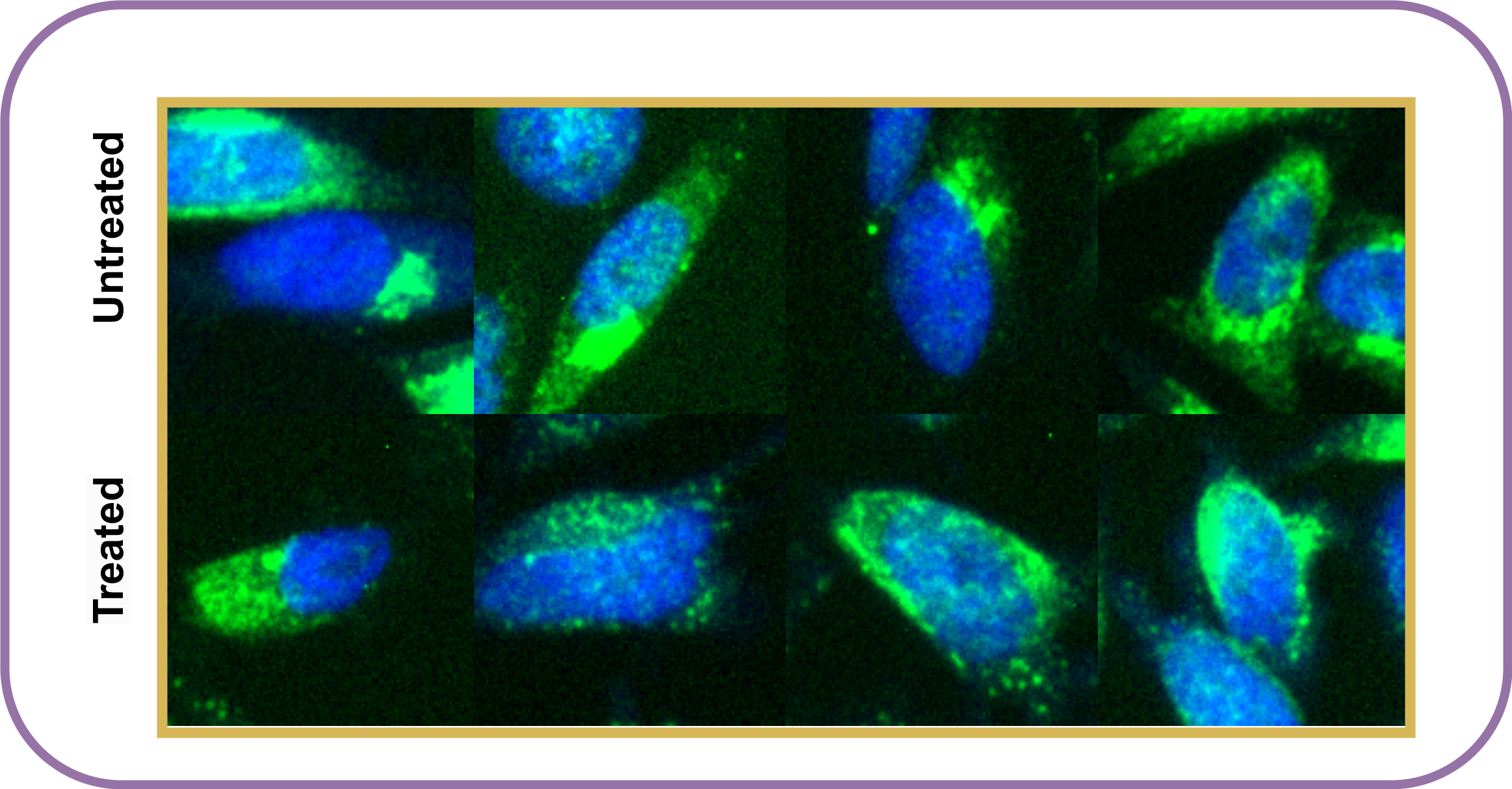} 
    \caption{}
    \label{fig:subfig2}
  \end{subfigure}

  \caption{Images from two datasets: \textbf{(a)} BBBC021 dataset and \textbf{(b)} Golgi dataset. While the differences between the two classes are apparent in BBBC021—such as the disappearance of the cytoplasm and fewer nuclei—they are more subtle in the Golgi dataset.}
  \label{fig:bio_dataset_stacked}
\end{figure}

Indeed, our goal is not only to generate images using this semantic code, but also to ensure that the generated image retains the same classification score as the original input. To achieve this, we introduce a classifier loss, which in our case is a KL divergence between the classification scores of the input image $x$ and the reconstructed one $x'$, an approach similar to~\cite{explaining_in_style}, the classifier loss is given by:
\begin{equation}
  \mathcal{L}_{\text{cls}} = D_{KL} \left[ C(x') \| C(x) \right]  
\end{equation}

The total loss to optimize is then: 

\begin{equation}
\mathcal{L}_{\text{sem}} = L_{\text{diffusion}} +\lambda_1 \mathcal{L}_{\text{cls}}
\end{equation}

where $\lambda_1$ is a hyperparameter.

\subsection{Finding interpretable directions in the latent space}
After training our semantic encoder, we introduce a contrastive learning approach to identify distinct and interpretable directions within its latent space. Contrastive learning has shown strong potential in exploring the latent spaces of GANs~\cite{latent_clr} and has been adapted recently to discover latent directions in the noise space of text-to-image diffusion models~\cite{noise_clr}. Unlike these prior methods, which locate semantic directions within either an intermediate GAN layer or the noise space of a diffusion model, our approach focuses on identifying meaningful directions directly within the latent space of the learned encoder.

Formally, given an inverted image noise  $x_T\in\mathcal{Z}_1$ and a semantic latent code $z_{sem}\in\mathcal{Z}_2$, we denote te diffusion models $\mathcal{DDIM} : \mathcal{Z}_1 \times \mathcal{Z}_2  \rightarrow \mathcal{X}$, where $\mathcal{X}$ is the space of images. We aim to find directions  $\Delta \mathbf{z}_1, \cdots, \Delta \mathbf{z}_N, \; N > 1$ such that for $k<N$,   $\mathcal{DDIM}(x_T, z_{sem}+\Delta \mathbf{z}_k)$ has visually meaningful changes compared to $\mathcal{DDIM}(x_T, z_{sem})$ while being similar to it. 

Specifically, we want to learn a mapping $\mathcal{D}_k : \mathcal{Z}_2 \times \mathbb{R} \rightarrow \mathcal{Z}_2$ that takes as input a latent code $z_{sem}$ and shift it along $\Delta \mathbf{z}_k$ with a weight $\alpha$, ie, $\mathcal{D}_k : (\mathbf{z}, \alpha) \rightarrow \mathbf{z} + \Delta \mathbf{z}_k$. Similar to~\cite{latent_clr}, we use multi-layer perceptron networks to learn the direction model $\mathcal{D}_k$ as follows:

\begin{equation}
\mathcal{D}_k(z, \alpha) = z + \alpha \frac{\mathcal{MLP}_1(z)}{\| \mathcal{MLP}_1(z)\|}
\end{equation}

For each latent code $z_i$, we shift it according to the $N_{th}$ directional models, as follows:
\begin{equation}
\mathbf{z}_i^k = \mathcal{D}(\mathbf{z}_i, \alpha)
\end{equation}

Then, we pass it through another MLP to obtain intermediate feature representations, 

\begin{equation}
\mathbf{h}_i^k = \mathcal{MLP}_2(\mathbf{z}_i, \alpha)
\end{equation}

After that, we compute the feature differences between the shifted and the original latent codes.

\begin{equation}
\mathbf{f}_i^k = \mathbf{h}_i^k -  \mathcal{MLP}_2(\mathbf{z}_i)
\end{equation}

Following contrastive learning principles, we aim to increase the similarity between edits originating from the same directional model, encouraging them to attract each other. Conversely, we want edits from different directional models to repel each other by reducing their similarity. This objective can be expressed by the following contrastive equation:

\begin{equation}
\ell_{cont}(z_i^k) = - \log \frac{\sum_{j=1}^N \mathbf{1}_{[j \neq i]} \exp\left(\operatorname{sim}(f_i^k, f_j^k) / \tau\right)}{\sum_{j=1}^N \sum_{l=1}^K \mathbf{1}_{[l \neq k]} \exp\left(\operatorname{sim}(f_i^k, f_j^l) / \tau\right)}
\end{equation}

The feature divergences obtained from the same directional model, represented as $\mathbf{f_1^k}, \mathbf{f_2^k}, \dots, \mathbf{f_N^k}$, are treated as \textit{positive pairs}. We aim to maximize their similarity, contributing to the numerator of the loss function. Conversely, feature divergences originating from different directional models (e.g., $\mathbf{f_1^k} \neq \mathbf{f_1^l}, , l \neq k$) are treated as \textit{negative pairs}. For these, we seek to minimize similarity, thus they contribute to the denominator of the loss function.

On top of the contrastive loss, we introduce a regularization term that promotes further decorrelation between the learned directions by minimizing the off-diagonal elements of the covariance matrix associated with the different directional models. This approach is inspired by~\cite{vicreg}, and the regularization term is defined as follows:
\begin{equation}
\mathcal{L}_{\text{reg}} = \sum_{i \neq j} \text{Cov}(\mathcal{D}_i(z), \mathcal{D}_j(z))^2
\end{equation}
Finally, we minimize this following total loss to learn the direction models:
\begin{equation}
\mathcal{L}_{\text{dir}} = L_{\text{cont}} +\lambda_2 \mathcal{L}_{\text{reg}}
\end{equation}

where $\lambda_1$ is a hyperparameter.

\subsection{Ranking the identified direction according to their importance}
After obtaining the directional models, the next step is to identify those that significantly influence the classifier’s probabilities. To do this, we first select a sample of images and compute their initial classification scores. For each discovered direction, we shift all images in the sample along that direction by a specific value of $\alpha$ and then calculate the new classification scores for the shifted images. If the average change in classification scores exceeds a predefined threshold, we retain that direction. Once a direction is selected, the images used to explain it are removed from the sample to avoid redundancy. This process is repeated iteratively until we identify the desired number of directions or exhaust the available images. The detailed pseudo-code for this procedure is provided in the Supplementary.~\ref{sec:algo}.

\begin{figure*}[ht]
  \centering
  \includegraphics[width=\linewidth, height=0.5\linewidth]{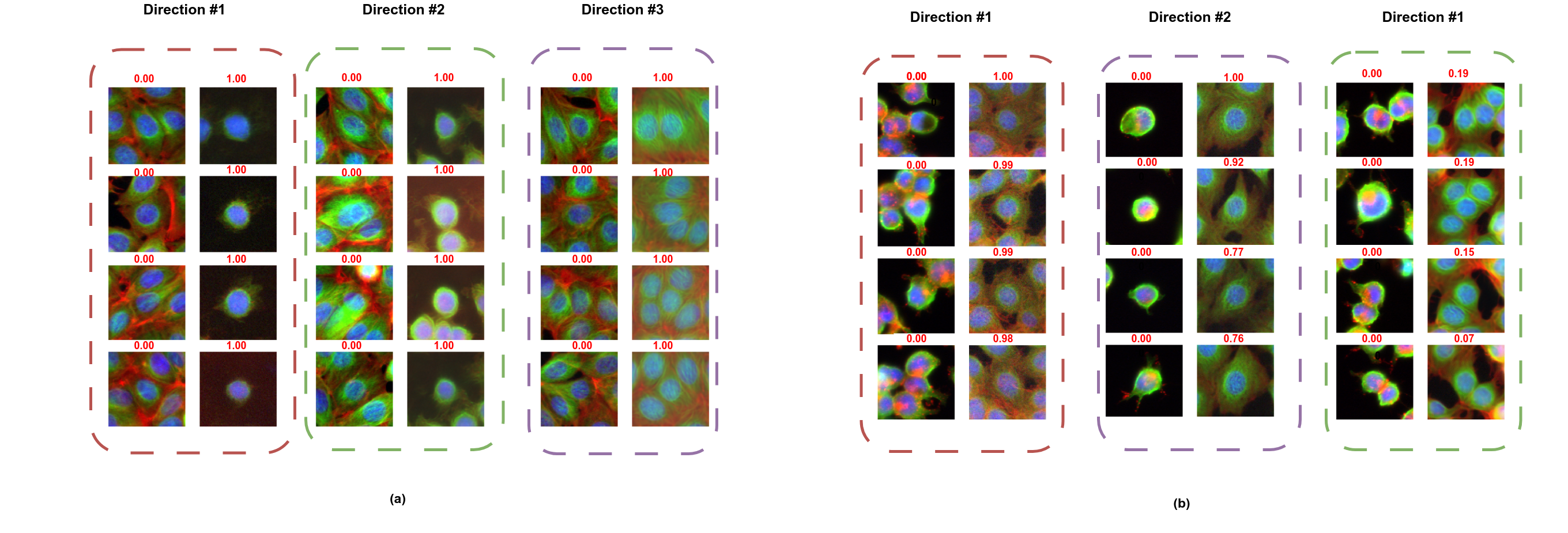} % Replace with your image file name
  \caption{Shifting images toward the opposite class. \textbf{Left}: DiffEx identified three distinct directions for transitioning from the untreated to the treated class. Direction 1 eliminates the cytoplasm and most cells, leaving a single nucleus at the center. Direction 2 removes the cytoplasm without eliminating all nuclei. Direction 3 tends to cluster nuclei closer together and decreases the intensity of the red channel. \textbf{Right}: To shift from the treated to the untreated class, Direction 1 increases the intensity of the red channel and pushes nuclei apart. Direction 2 enhances the green channel, while Direction 3 increases the cell count, replicating known phenotypes}
  \label{fig:bio_figure}
\end{figure*}

\section{Results}

\subsection{Datasets}

We used the following datasets to evaluate our method:

\begin{figure}[ht]
  \centering
  \includegraphics[width=\linewidth]{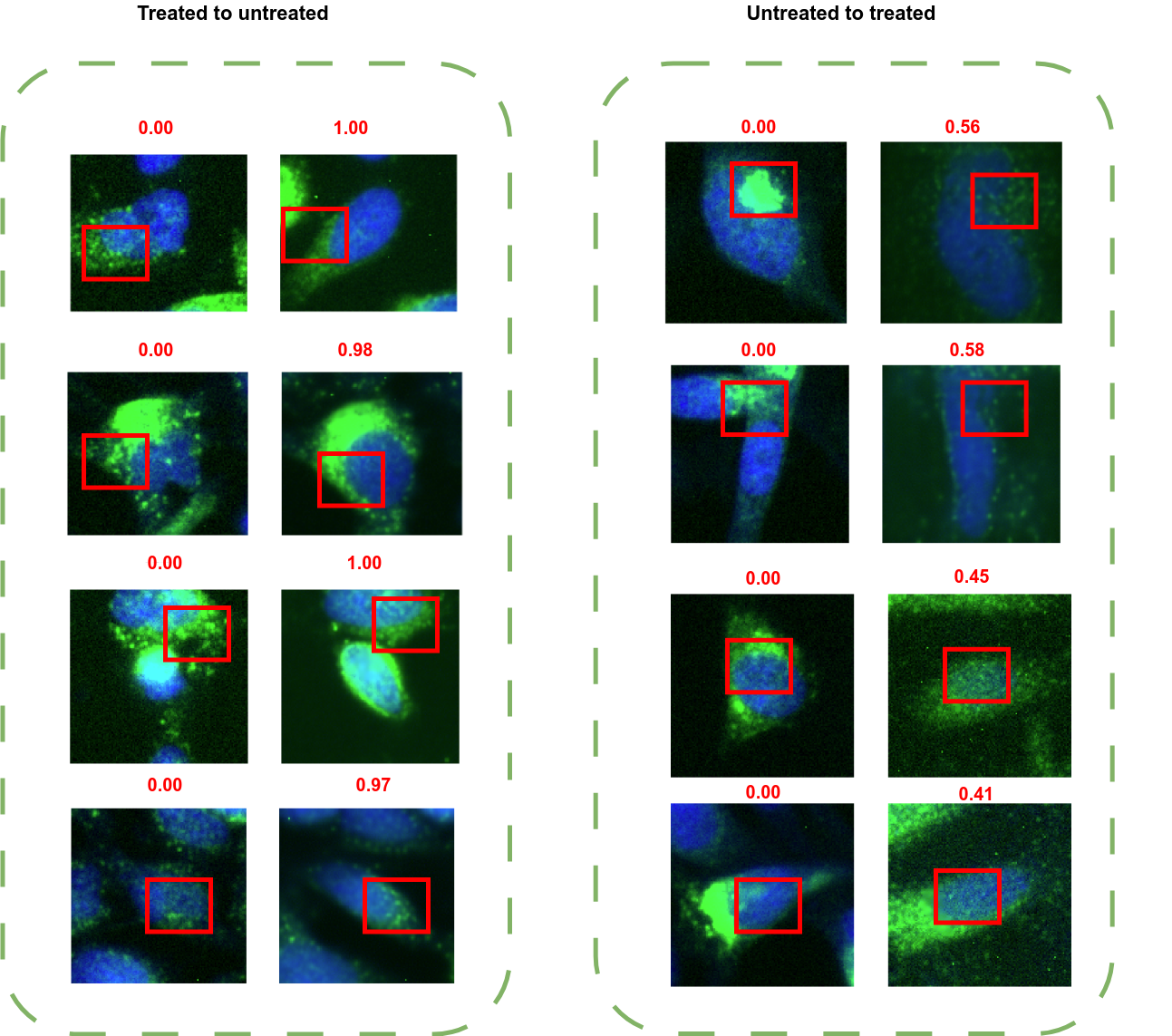} % Replace with your image file name
  \caption{Shifting images toward the opposite class. \textbf{Left}: When transitioning from the treated to the untreated class, the Golgi apparatus tends to aggregate. \textbf{Right}: Conversely, shifting from the untreated to the treated class results in its dispersion. These observations replicate the phenotypic effects of the treatment, which induces Golgi apparatus scattering.}
  \label{fig:figure_golgi_first}
\end{figure}

\noindent \textbf{FFHQ}:
The FFHQ~\cite{stylegan} dataset is a high-quality image collection containing 70,000 high-resolution face images with diverse variations. Given its combination of high resolution and diversity, FFHQ has become a benchmark in the field.

\noindent \textbf{BBBC021}:
The BBBC021 dataset \cite{bbbc} is a publicly available collection of fluorescent microscopy images of MCF-7, a breast cancer cell line treated with 113 small molecules at eight different concentrations. For our research, we focused on images of untreated cells and cells treated with the highest concentration of the compound Latrunculin B. In Fig.~\ref{fig:bio_dataset_stacked}, the green, blue, and red channels label for B-tubulin, DNA, and F-actin respectively.

\noindent \textbf{Golgi}:
Fluorescent microscopy images of HeLa cells untreated (DMSO) and treated with Nocodazole. In Fig. \ref{fig:bio_dataset_stacked}, the green and blue channels label for B-tubulin and DNA respectively.

\subsection{DiffEx encodes natural and biological images}

We trained a classifier on the FFHQ dataset to distinguish between male and female classes, we also trained classifiers on BBBC021 and Golgi datasets to classify untreated and treated images. As shown in Table~\ref{tab:metrics_comparison}, the proposed framework effectively encodes both biological and natural image features. Indeed, the different metrics used to assess the reconstruction quality demonstrate very low values for the datasets utilized in the experiments. Furthermore, the classification metrics across the three classifiers perform well on the generated images.
This consistent classification accuracy suggests that the generated images are not only visually coherent but also maintain key distinguishing features necessary for correct classification, most importantly, the absence of adversarial artifacts that could alter the classifier's decisions.

\begin{table}[h]
\centering
\caption{Comparison of metrics for different datasets, including classification accuracy.}
\begin{tabular}{lcccc}
\toprule
Dataset          & LPIPS   & SSIM   & MSE    & Accuracy \\
\midrule
BBC021           & 0.0237  & 0.99   & 0.0007 & 100 \\
FFHQ/gender      & 0.0118  & 1.0    & 0.0004 & 99.5 \\
Golgi            & 0.0594  & 1.0    & 0.0003 & 95 \\
\bottomrule
\end{tabular}
\label{tab:metrics_comparison}
\end{table}

\subsection{Explaining a Classifier trained on natural and biological images}

First, we applied DiffEx to explain a classifier trained on natural images. In Fig.~\ref{fig:figure_faces}, some directions identified by the method on the FFHQ dataset are shown. Specifically, short haircuts tend to push the classification toward the "male" class, while the presence of lipstick pushes the classification toward the "female" class, more examples are shown in Supplementary.~\ref{sec:more_examples}

We then applied DiffEx to a classifier trained on the BBBC021 images. In Fig.\ref{fig:bio_figure}, we illustrate the three most significant directions identified by our method for transitioning between the treated and untreated cases. Each direction leads to distinct outputs, demonstrating that the directions are well disentangled and separated. These directions replicate various phenotypic aspects induced by the treatment administered to the cells. As shown in Fig.\ref{fig:bio_dataset_stacked}, the drug’s toxicity causes cell death, leading to the disappearance of cytoplasm and a reduction in nuclei count. Direction 1 replicates this phenotype: the generated image displays only a single nucleus centered in the frame, with the cytoplasm entirely absent. In contrast, Direction 2 does not entirely eliminate the cells but removes most of the cytoplasm, a hallmark of the treatment effect. Direction 3 maintains the cell count and partially retains the cytoplasm but reduces the intensity of the red channel, it also tends to cluster the nuclei closer together. 

For the reverse case shown in Fig.~\ref{fig:bio_figure}, directions were identified for transitioning from the treated to the untreated class. Direction 1 adds cytoplasm back and increases the distance between nuclei, effectively reversing the phenotype of Direction 3 from the untreated-to-treated transition. Direction 2 restores cytoplasm while keeping the nuclei count constant. Finally, Direction 3 increases the number of nuclei and slightly restores cytoplasm between them.

Lastly, we tested our method on another dataset, the Golgi dataset. These images depict cells treated with Nocodazole, which causes the Golgi apparatus to scatter. This phenotype can be subtle and challenging to observe. Using CellProfiler, we confirmed this phenotype by measuring the area occupied by the Golgi apparatus in both untreated and treated cases. As shown in Fig., the Golgi apparatus occupies a larger area in the untreated case due to its scattering.

In Fig.~\ref{fig:figure_golgi_first}, we highlight the most significant direction identified by the method. For the untreated-to-treated transition, the Golgi apparatus becomes more scattered, replicating the effect of the treatment. Conversely, for the treated-to-untreated transition, the Golgi apparatus becomes more aggregated, effectively mimicking the reversal of the treatment’s effect. In contrast to the BBBC021 dataset, all the identified directions replicate exactly the same phenotypes. This could be due to the limited number of channels used in this dataset (green and blue only).

\begin{figure}[h]
  \centering
  \includegraphics[width=\linewidth]{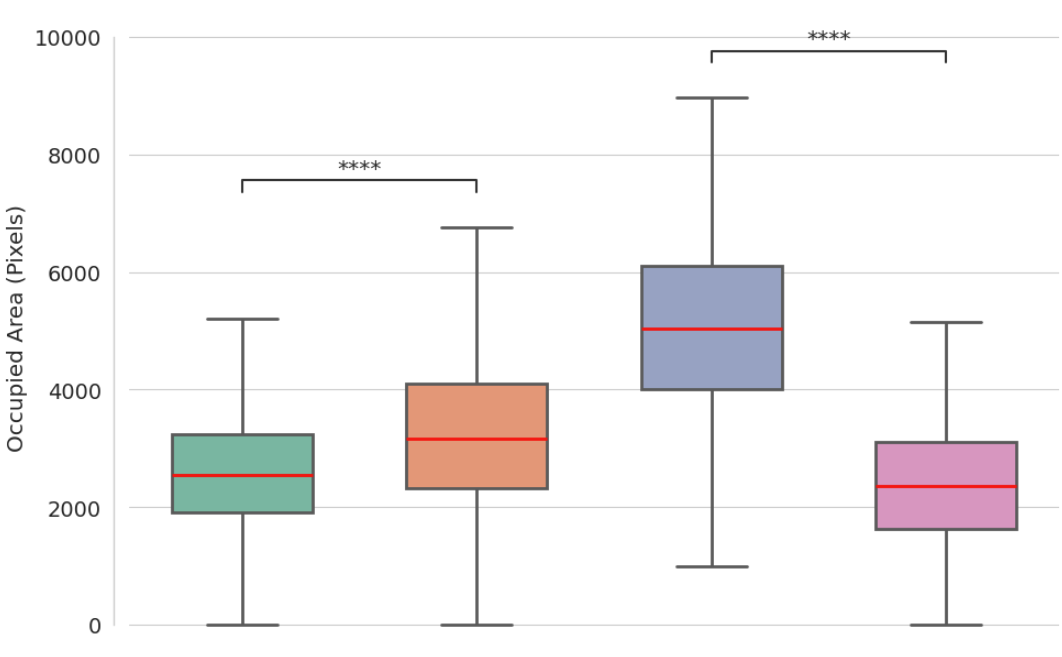} % Replace with your image file name
  \caption{\textbf{Left}: Measurement of the Golgi apparatus area in real images for both conditions reveals a difference in its spatial distribution. The area is larger in the treated case due to treatment-induced scattering. \textbf{Right}: Measurement of the nuclear area in the BBBC021 dataset shows that it is larger in the untreated case. This is attributed to the treatment’s toxicity, which eliminates cells, reducing overall nuclear presence}
  \label{fig:boxplots}
\end{figure}

\begin{table*}[htbp]
\centering
\caption{Performance metrics of our method compared to GCD. A '-' indicates that no explanation was found}
\label{tab:performance_metrics_fid}
\begin{tabular}{l|cc|cc|cc}
\toprule
\multirow{2}{*}{\textbf{Method}} 
& \multicolumn{2}{c|}{\textbf{Gender}} 
& \multicolumn{2}{c|}{\textbf{BBBC021}} 
& \multicolumn{2}{c}{\textbf{Golgi}} \\
\cline{2-7}
& \textbf{KID} & \textbf{SSIM} 
& \textbf{KID} & \textbf{SSIM} 
& \textbf{KID} & \textbf{SSIM} \\
\midrule
GCD  & 0.13 & 0.55 & -    & -    & -      & -    \\
Ours & \textbf{0.12} & \textbf{0.67} & 0.07 & \textbf{0.22} & \textbf{0.032} & \textbf{0.69} \\
\bottomrule
\end{tabular}
\end{table*}

\subsection{Comparing to existing methods}

Comparing our method to existing approaches is inherently challenging, as many of the current methods for detecting phenotypes rely solely on generative models. Among the most closely related methods, $GCD$\cite{global_counterfactual_explanations} stands out, although it was not proposed to identify phenotypes, it uses diffusion models to explain a classifier. Similar to our approach, they utilize a latent space constructed with DiffAE\cite{diffusion_ae}. However, $GCD$ does not incorporate the classifier during training, and it identifies counterfactual directions using a single image optimized to minimize a counterfactual loss.

\begin{figure}[h]
  \centering
  \includegraphics[width=0.8\linewidth]{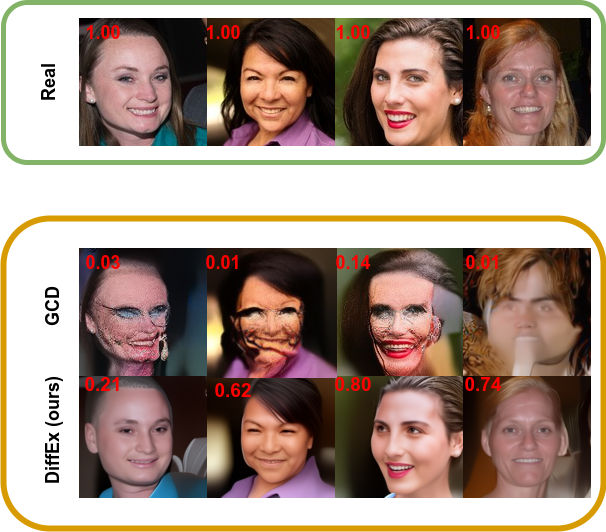} % Replace with your image file name
  \caption{Generating counterfactual explanation with our method and GCD. We can see that our method gives visually better and more disentangled results. }
  \label{fig:figure_ffhq}
  
\end{figure}

\begin{figure}[h]
  \centering
  \includegraphics[width=0.8\linewidth]{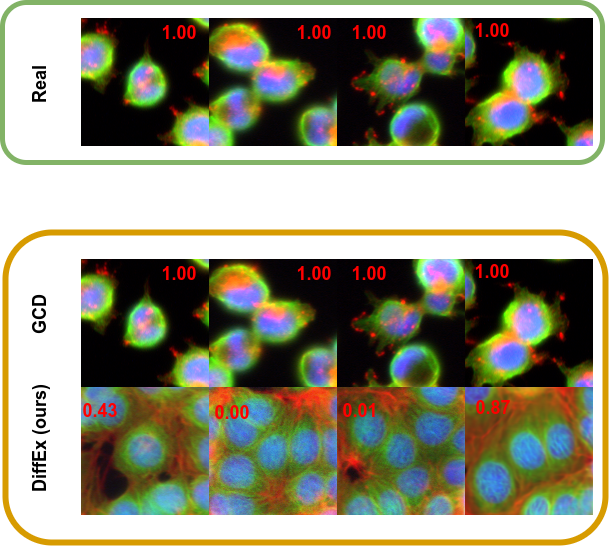} % Replace with your image file name
  \caption{Generating counterfactual explanations with our method and GCD reveals that GCD fails to generate counterfactuals for biological datasets. This limitation may be due to its reliance on a single image to identify directions in the latent space, which proves challenging for datasets with high variability, such as biological data.}
  \label{fig:figure_golgi}
\end{figure}
For comparison, we identified the first principal direction that most significantly shifts the classification score of the trained classifier. In Fig.~\ref{fig:figure_ffhq}, we present the generated explanation using our method and GCD. It is evident that the explanations produced by our method are visually superior and more disentangled compared to those obtained using GCD. Specifically, our method focus on modifying a single attribute—primarily shortening the hairstyle—while GCD introduces changes to multiple attributes simultaneously, leading to less interpretable results. Additionally, we observe that the classification shifts are more pronounced in the examples generated by GCD compared to those produced by our method. This can be attributed to GCD's optimization of the counterfactual loss with respect to shifts in latent space. While GCD can identify a direction that reduces the classifier's confidence, the resulting counterfactuals are often of poor visual quality, as evident in some of the generated samples. In Fig.~\ref{fig:bio_figure}, we further evaluate the performance of GCD and our method on biological images. Notably, GCD fails to generate meaningful images when applied to this domain. This limitation is likely due to GCD's reliance on a single image to identify directions in the learned latent space. While this approach works well in datasets with inherent class similarities, such as FFHQ, it struggles in scenarios where there is high variability between classes, as is the case with biological images. Furthermore, in Table~\ref{tab:performance_metrics_fid}, we compare the quality of the generated explanations using the Kernel Inception Distance (KID)~\cite{kid}, as well as the similarity between the original and generated images. The results show that our method consistently outperforms GCD across various datasets. This indicates that our method produces images that are not only closer to the target dataset distribution but also retain higher similarity to the original images, demonstrating its effectiveness and robustness.

\section{Conclusion}

In this work, we introduced DiffEx, a versatile framework for explaining classifiers using diffusion models. By identifying meaningful directions in the latent space, DiffEx produces high-quality and disentangled attributes that maintain fidelity to the original data while effectively shifting classification outcomes. An important application of DiffEx is its ability to detect phenotypes. We validated this capability across multiple datasets, demonstrating that DiffEx can reveal fine-grained biological variations and enhance our understanding of cellular and phenotypic differences. This highlights the method's potential to be a valuable tool in advancing research in biology and related fields, where uncovering subtle variations is essential. Moreover, DiffEx can be extended to other applications where it is critical to explain classifier outputs, making it a versatile framework for enhancing model interpretability across diverse domains.

\bibliography{example_paper}
\bibliographystyle{icml2025}

%%%%%%%%%%%%%%%%%%%%%%%%%%%%%%%%%%%%%%%%%%%%%%%%%%%%%%%%%%%%%%%%%%%%%%%%%%%%%%%
%%%%%%%%%%%%%%%%%%%%%%%%%%%%%%%%%%%%%%%%%%%%%%%%%%%%%%%%%%%%%%%%%%%%%%%%%%%%%%%
% APPENDIX
%%%%%%%%%%%%%%%%%%%%%%%%%%%%%%%%%%%%%%%%%%%%%%%%%%%%%%%%%%%%%%%%%%%%%%%%%%%%%%%
%%%%%%%%%%%%%%%%%%%%%%%%%%%%%%%%%%%%%%%%%%%%%%%%%%%%%%%%%%%%%%%%%%%%%%%%%%%%%%%
\newpage
\appendix
\onecolumn

\section{More examples}
\label{sec:more_examples}

In the following examples, we trained DiffeEX to identify 10 different directions in the semantic space. As we can see, these directions alter various attributes, but not all of them lead to changes in the output probabilities. To address this, we apply our ranking algorithm to rank the directions based on their ability to modify the classification output. For instance, in this case, the most important attribute is direction 5 (positive), which shortens the haircut of images belonging to the female class. Conversely, direction 6 (negative) adds makeup to images of males, increasing the probability of classification into the female class.

% First figure

\begin{figure}[H]  % [H] forces the figure to be placed right here
  \centering
  \includegraphics[width=\linewidth]{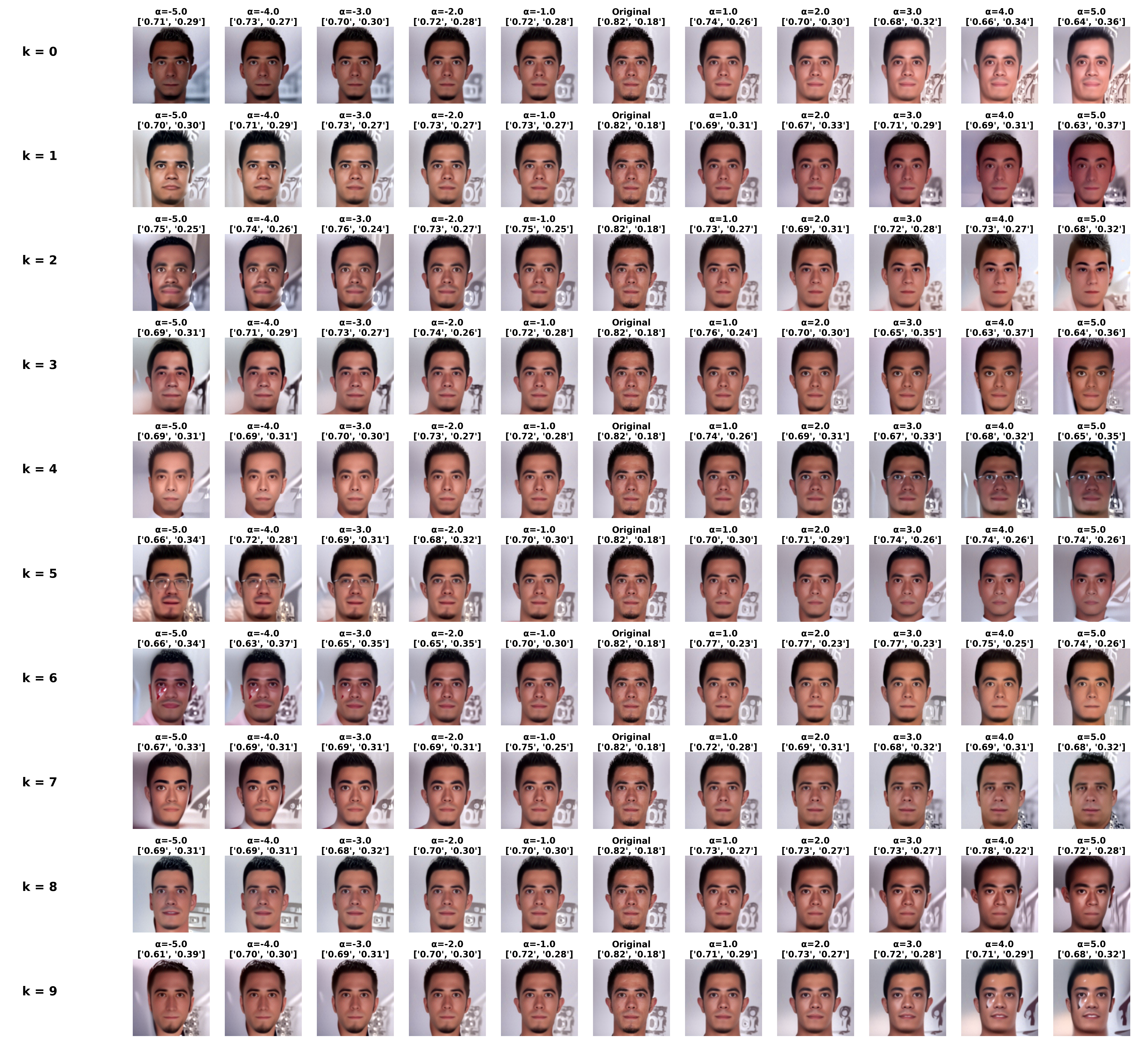}
  \label{fig:sample_2}
\end{figure}

%%%%%%%%%%%%%%%%%%%%%%%%%%%%%%%%%%%%%%%%%%%%%%%%%%%%%%%%%%%%%%%%%%%%%%%%%%%%%%%
%%%%%%%%%%%%%%%%%%%%%%%%%%%%%%%%%%%%%%%%%%%%%%%%%%%%%%%%%%%%%%%%%%%%%%%%%%%%%%%

\newpage
\section{Ranking Algorithm Pseudo-code}
\label{sec:algo}

\begin{figure*}[ht]
  \centering
  \includegraphics[width=0.9\linewidth]{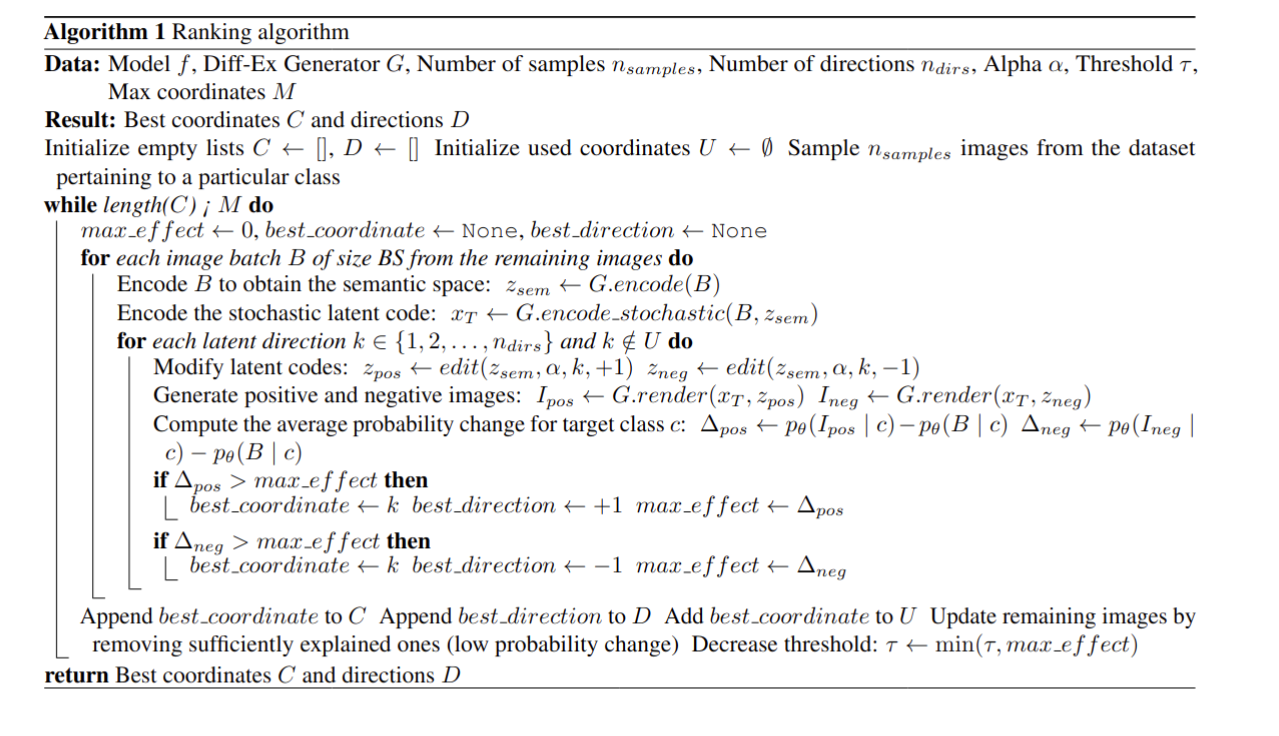} % Replace with your image file name
\end{figure*}

\end{document}